\documentclass[10pt,twocolumn,letterpaper]{article}
\usepackage{graphicx}
\usepackage{amsmath}
\usepackage{hhline}
\usepackage{geometry}
\usepackage{booktabs}
\usepackage{arydshln}
\usepackage{amsfonts}
\usepackage[pagenumbers]{cvpr} 

\usepackage[dvipsnames]{xcolor}

\definecolor{cvprblue}{rgb}{0.21,0.49,0.74}
\usepackage[pagebackref,breaklinks,colorlinks,citecolor=cvprblue]{hyperref}

\def\paperID{*****} 
\def\confName{CVPR}
\def\confYear{2024}

\title{Progressive Alignment with VLM-LLM Feature to Augment \\Defect Classification for the ASE Dataset}

\author{$^\dagger$$^1$Chih-Chung Hsu, $^\ddagger$$^1$Chia-Ming Lee, $^2$Chun-Hung Sun, $^2$Kuang-Ming Wu\\
$^1$Institute of Data Science, National Cheng Kung University, Taiwan\\
$^2$Corporate R\&D, Advanced Semiconductor Engineering Group, Taiwan\\
{\tt\small $^\dagger$cchsu@gs.ncku.edu.tw, $^\ddagger$zuw408421476@gmail.com}
}

\begin{document}
\maketitle
\begin{abstract}

%
Traditional defect classification approaches are facing with two barriers. 
\textbf{(1) Insufficient training data and unstable data quality.} Collecting sufficient defective sample is expensive and time-costing, consequently leading to dataset variance. It introduces the difficulty on recognition and learning. 
\textbf{(2) Over-dependence on visual modality.} When the image pattern and texture is monotonic for all defect classes in a given dataset, the performance of conventional AOI system cannot be guaranteed.
%
%
A main question is, \textbf{"how to solve those two problems when they occur at the same time?"} The feasible strategy is to explore another feature within dataset and combine an eminent vision-language model (VLM) and Large-Language model (LLM) with their astonishing zero-shot capability. 
In this work, we propose the special ASE dataset, including rich data description recorded on image, for defect classification, but the defect feature is uneasy to learn directly. 
Secondly, We present the prompting for VLM-LLM against defect classification with the proposed ASE dataset to activate extra-modality feature from images to enhance performance. Then, We design the novel progressive feature alignment (PFA) block to refine image-text feature to alleviate the difficulty of alignment under few-shot scenario. 
Finally, the proposed Cross-modality attention fusion (CMAF) module can effectively fuse different modality feature.
Experiment results have demonstrated our method's effectiveness over several defect classification methods for the ASE dataset.
\end{abstract}    
\section{Introduction}
\label{sec:intro}

\begin{figure}
    \begin{center}
    \includegraphics[width=0.42\textwidth]{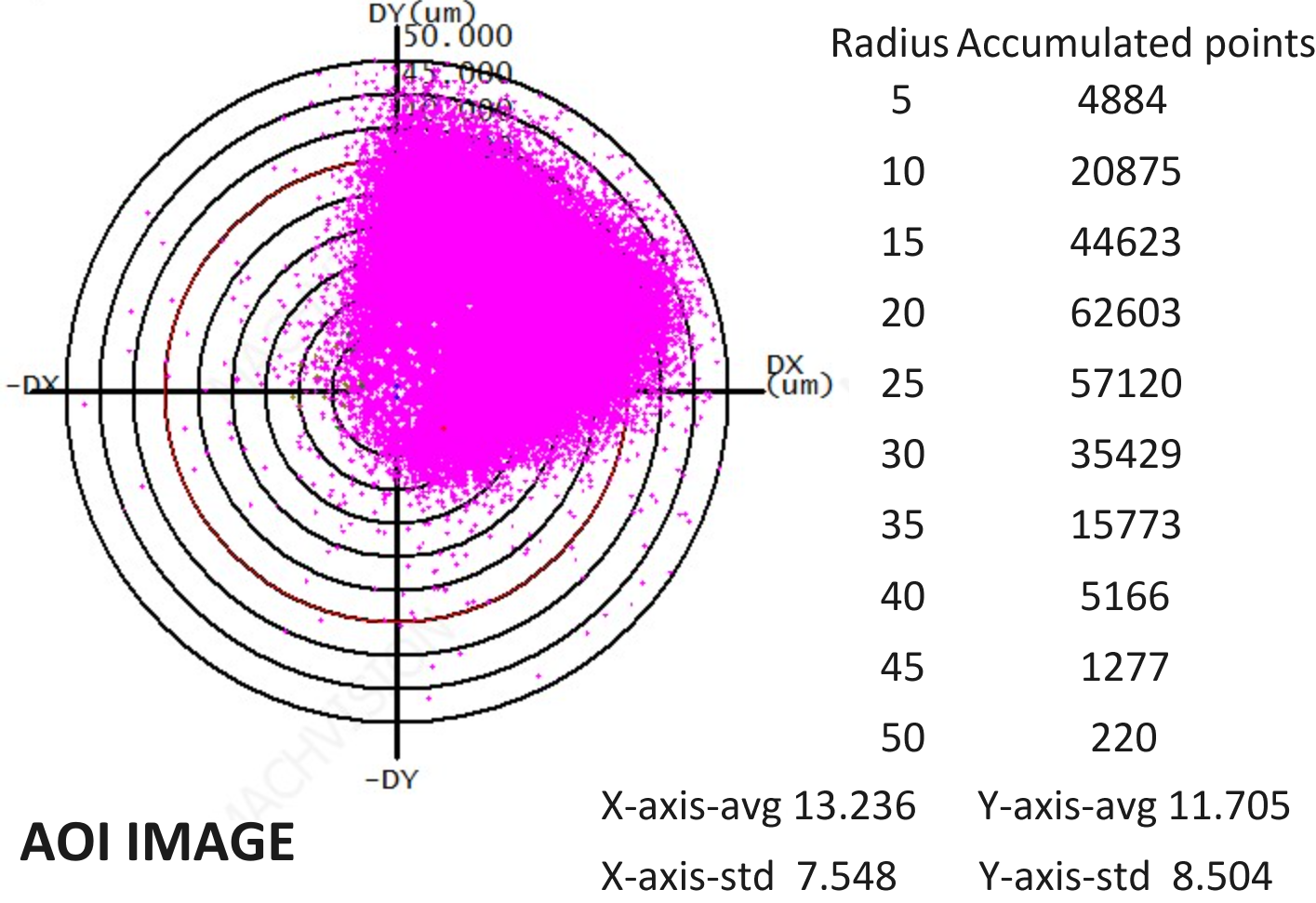}
    \end{center}
    \caption{The brief illustration of proposed ASE data. It contains two parts: (1) AOI image, which formed by the large number of pink dot. (2) The recorded numeric and textual information corresponds to the pink-dot.}
     \label{ASEexample.png}
\end{figure}
In the realm of industrial manufacturing, defect recognition is paramount, serving a dual purpose: enhancing product quality and efficiency and indirectly curbing production costs by reducing the prevalence of both false negatives and positives. At its core, defect detection leverages a myriad of analytical instruments to discern features rooted in the physical properties of diverse products. Once these features are harvested, models are then deployed for defect recognition.

Among the plethora of defect classification strategies, Automatic Optical Inspection (AOI) emerges as a predominant choice, especially in the industrial domain. The cost of manual inspection is high, and accuracy may decrease due to human fatigue. AOI harnesses the power of high-resolution imaging tools to inspect products throughout the manufacturing phase. It taps into deep-learning approach like convolutional neural network (CNN) in recent years, to spot defects. By juxtaposing the visual attributes of standard samples with defective ones, model can efficiently pinpoint anomalies. Many researchers have proposed several defect detect or classification method to improve the quality of system, such as \cite{RudWeh2022,detect_survey,detect_survey_2,detect_survey_3,Haurum_2021_CVPR, traditional1,traditional2,paperwithcodesota1,paperwithcodesota2}.

Despite its success over the years, however, the AOI methodology isn't without its limitations. For example, datasets are of low quality and have insufficient sample sizes, which may be due to the high cost of collecting defect data. Unpredictable lighting conditions and camera shifts within AOI system undercut the reliability of model. Even more challenges like sparse training data and sample imbalances can yield performance that falls short of expectations.These crucial factors result in the degradation of deep model. Several work utilize ensemble method \cite{fewshot_defect_1} or feature enhancement \cite{fewshot_defect_2,fewshot_defect_3} approach to improve performance, the effect is still limited due to changes in different data sets and various negative factors.

On the other hand, the dawn of multimodal learning has ushered in a new era where features from disparate modalities can be synergized, amplifying model performance. This integration has catalyzed advancements in domains between computer vision and natural language processing. Many researcher have proposed cross-modality learning framework, boosting the downstream application by using the well-known Vision-Language Model (VLM) and Large Language Model (LLM). Its influence spans the field of deep learning. Due to its superiority of zero-shot capablilities learned from numerous and high quality image/text data pairs, there is a possible way to augment defect classification system by extracting VLM-LLM feature without cost-consuming data collection and laborious labeling, further solving the data insufficiency and over-dependence on AOI and vision modality.

The major novelties and contributions of this paper can be divided three-folds as follow:

\begin{itemize}
    \item \textbf{Prompting with VLM-LLM to augment performance for the proposed ASE dataset}: Traditional vision-based methods cannot easily solve the problem of ASE dataset. (1) The pattern of ASE data is very monotonous, and (2) the number of samples is insufficient. We leverage the zero-shot capabilities of VLM-LLM through prompting engineering to capture external-modal features to improve the performance on binary and multi-class classification task for the ASE dataset.

\item \textbf{Progressive feature alignment block}: the novel Progressive Feature Alignment (PFA) block effectively aligns image-text representation with Progressive Training Strategy (PTS) and contrastive learning manner, and selects the negative samples at beginning. Afterwards, we gradually samples more training data to align features between two image-text data pairs iteratively, addressing the challenge of features being difficult to align in a small number of samples.
    
\item \textbf{Cross-modality attention fusion module and Task-specific Data Augmentation}: the proposed Cross-Modality Attention Fusion (CMAF) module enables our model to adaptively fuse the high fidelity features from different modality branches. With the dedicated Task-specific Data Augmentation (TDA) for the ASE dataset, the source domain can be enlarged, further improving the recognition ability against novel samples. Experiment results have demonstrated our method's promising performance compared with other methods for the ASE dataset.

\end{itemize}

\section{Related Work}
\label{sec:relatedwork}

\begin{figure*}
\includegraphics[width=\textwidth]{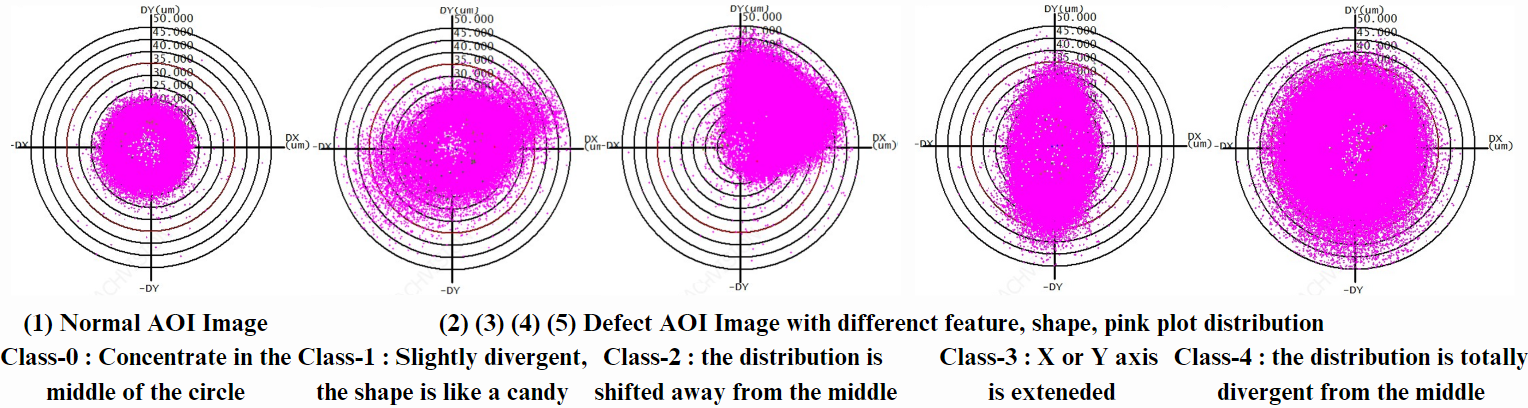}
\caption{The brief summary of different classes within ASE dataset. The dense pink dots in the picture form a special pattern. This pattern can be described by VLM easily, and we can further combine it with LLM by suitable prompting engineering and our prior knowledge on ASE dataset to enhance the defect classification performance without any expensive fine-tuning for VLM or LLM.}
 \label{asecategory.png}
\end{figure*}

\subsection{Defect Recognition under Few-shot Scenarios}

Defect Recognition has witnessed remarkable enhancements with the integration of deep and machine learning algorithms. In overall, defect recognition can be toward divided into two groups: (1) defect detection and (2) defect classification. The major issue of defect recognition is that when the dataset is small or the quality of image is disappointing, the performance of deep model will be limited.

\textbf{Defect Detection.} Generally speaking, defect detection aims to point out where the defect exist. Defect detection tasks are more common than classification. There have been various works and open-sources specifically designed to handle different types of defect detection, such as \cite{Surface-Defect-Detection,Anomalib,detect_survey,detect_survey_2,detect_survey_3}. Recently, in order to deal with the challenges in the above mentioned, CS-Flow \cite{RudWeh2022} leverage cross-scale feature flow to integrate the high and low-level feature, enhancing learned representation. XDNet \cite{fewshot_defect_3} is based on self-supervised learning strategy to explore the information within given dataset. 

\textbf{Defect Classification.} Different from detection, the goal of defect classification is to figure out which class does sample belong to. It can be seen as image classification task. The simple way is to train a CNN \cite{vgg,resnet,efficientnet} with suitable augmentation pipeline \cite{Shanmugam_2021_ICCV} or data synthesis \cite{PreAugNet}, and ensembling different CNN backbone to enhance the few-shot capabilities \cite{fewshot_defect_1}. Some researchers have proposed the incorporation of low-level features such as SIFT \cite{sift} or depth information \cite{ocr1,ocr2,ocr3} to bolster the model performance.

\subsection{Prompting Learning for VLM and LLM}

While traditional deep learning typically relies on single-modality data, CLIP \cite{CLIP} have pioneered the integration of vision-language. Benefit from the large-scale of high quality pre-training on vision-language data, multi-modality-based model \cite{ALBEF,BLIP,BLIP-2,InstructBLIP2,eva} have shown astonishing zero-shot capabilities. 
The VLM paradigm introduced by CLIP, along with the versatility of LLM like GPT-3 \cite{gpt3} and LLaMA \cite{llama,llama2}, have a significant impact on the deep learning community. VPT \cite{jia2022vpt} further enhance downstream task performance using fine-tuning-free prompt techniques.

\textbf{VLM Prompting in Vision} With the famous CoCoOp and MaPLe \cite{zhou2022coop,zhou2022cocoop, MaPLe}, which proposed prompting learning with VLM in vision, there has been a proliferation of works utilizing VLM to enhance computer vision task. Hu and Awal et al. utilized VLM-feature and accurate prompting to align the depth and visual semantic representation for few-shot depth estimation \cite{adapt1,adapt2}. Liang et al. proposed CLIP-LIT \cite{liang2023iterative}, which deployed iterative prompting strategy with CLIP prior for open-world image enhancement. Subramanyam et al., presented CREPE \cite{CREPE} and explored the potential of VLM to improve the performance of visual-object-relationship prediction.

\textbf{LLM Prompting in Vision} The concept of language instruction has also been embraced by the computer vision community for defining image-to-text tasks. Flamingo \cite{flamingo} stands out as a groundbreaking work in this regard, utilizing both vision and language inputs as prompts and achieving impressive few-shot results across various vision-language tasks like image captioning and Visual Question Answering (VQA) \cite{vqa,InstructBLIP2,BLIP,BLIP-2}. Prophet \cite{prophet1,prophet2} is designed to guide LLM using answer heuristics for knowledge-based VQA without incorporating external knowledge. VisionLLM \cite{visionllm}, presented remarkable results on several benchmarks.

\subsection{Multi-Modality Feature Alignment and Fusion}
Multi-modality alignment and fusion are both the most essential topics in multi-modal learning, which mainly focus on how to incorporate modality-wise features into a joint representation for downstream application. Different modality feature which embedded on high-dimensional space may contains complementarity, its information may benefits each other \cite{Adaptive}. Originated from the limitation of single-modality approaches, which suffer from the variation of dataset and their poor generalizablilty, CLIP \cite{CLIP} and ALBEF \cite{ALBEF}, which aim to align image-text pairs representation embedded in high-dimensional latent space, has been introduced, lighting up an alternative avenue for tackling zero/few-shot and low-quality dataset scenarios \cite{qu2022transmef,linmultimodality}. If multi-modal feature can be processed effectively, rich feature representation can be obtained \cite{qu2022transmef,Chitta2023PAMI,Prakash2021CVPR,10132374,D-LLM,gbacm}, outperforming the single-modality-based methods. 

Multi-modal fusion strategy can be typically classified into feature-level fusion, decision-level fusion and hybrid fusion \cite{surfey,surfey2,surfey3,surfey4}. Feature-level feature aims to conjuncte the different modality feature to archive high-level joint representation within single model \cite{wang2020cen,wang2022cenpami}. Decision-level fusion like the classical Mixture-of Experts (MoE) \cite{Moe} and model-ensemble strategy, each expert specializes in a subset of all given modalities. Hybrid fusion is the most complicated design \cite{xue2022dynamic,hf}, this type of strategy achieves the best results but has the highest computational burden.




\section{Methodology}
\label{sec:method}
\begin{figure*}
\includegraphics[width=\textwidth]{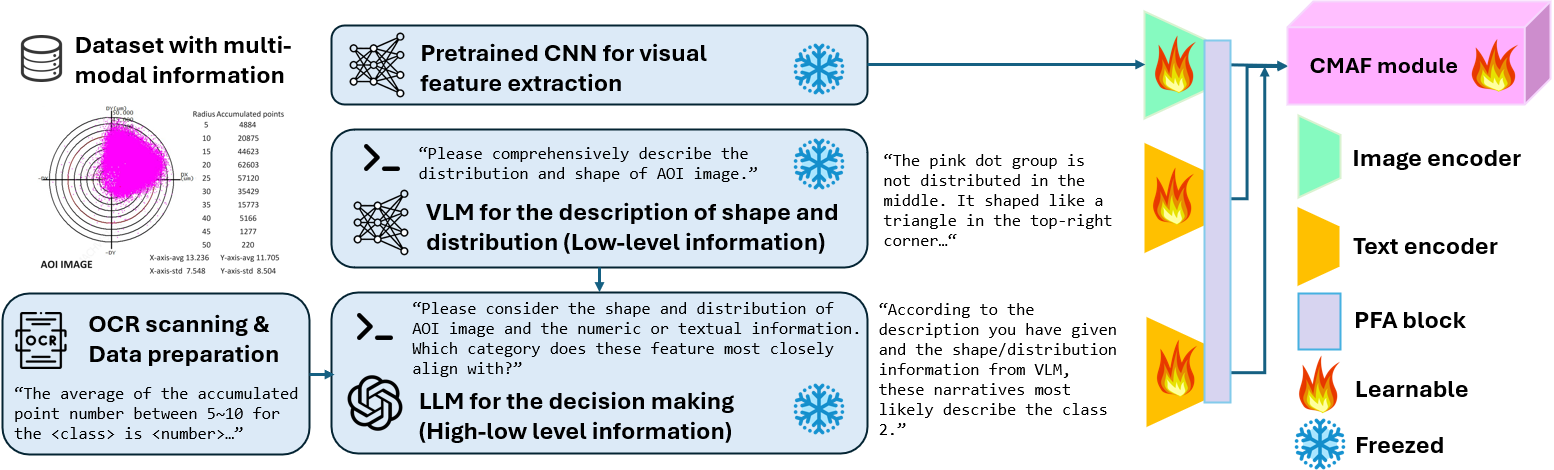}
\caption{The overall architecture of the proposed framework. It aims to incorporate with VLM-LLM to explore external-modality features to jointly learn better representations for defect classification. Through our proposed Progressive Feature Alignment (PFA) and Cross-Modality Attention Fusion (CMAF) module, textual and visual features are efficiently fused, effectively addressing the challenges and limitations commonly encountered by conventional deep learning approaches (e.g. CNN, ViT) when processing the ASE dataset.}
 \label{model.png}
\end{figure*}

An overview of our approach is shown in the Figure \ref{model.png}. We first introduce the ASE dataset and the limitation of conventional methods. Then, we illustrate our method in details.

\begin{figure}
    \begin{center}
    \includegraphics[width=0.48\textwidth]{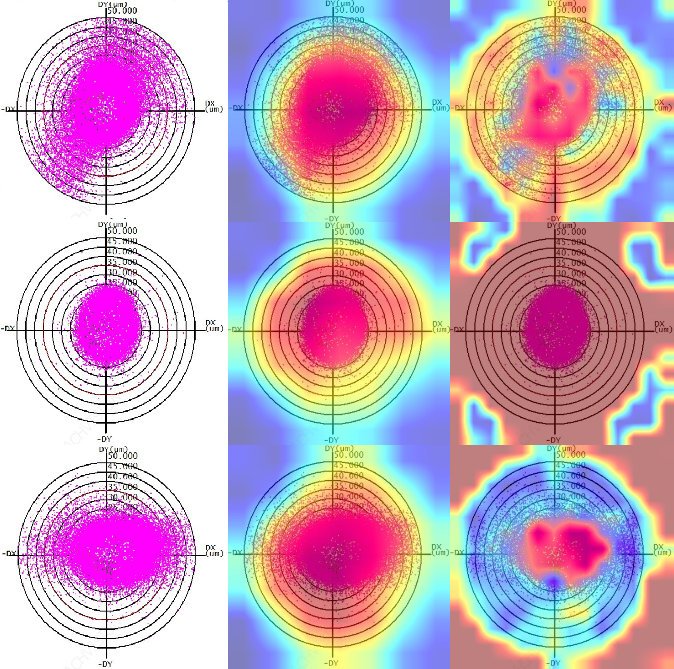}
    \end{center}
    \caption{The GradCAM++ \cite{8354201} visualization for proposed ASE dataset. Left column: AOI images from ASE dataset; Middle column: using ResNet50 \cite{resnet}; Right column: using DeiT \cite{touvron2020training}.}
     \label{gradcam.png}
\end{figure}

\subsection{Proposed ASE Dataset} The dataset is provided by ASE corporation and consists of two parts, the details is shown in Figure \ref{ASEexample.png}. This dataset contains five classes, with detailed insights provided in Figure \ref{asecategory.png}. (1) Image, which records the drilling machine's drilling within a fixed range. The goal is to ensure that the drilled holes are positioned as close to the center of the circle as possible, with no shape deviation (without defect). (2) The information about the drilled hole positions, including frequency statistics within specific intervals, as well as corresponding mean $\mu$ and standard deviation $\sigma$. The second part is recorded by meticulous machine, ensuring high quality.

\textbf{Limitations and Challenges.}  The ASE dataset itself exhibits a monotonous texture/pattern compared to conventional images. Due to its inductive biases, such as locality and spatial invariance, CNNs model can not learn sufficiently strong low-level feature within shallow layers. As for Vision Transformer (ViT) \cite{touvron2020training} approaches, while effectively focusing on the global dependencies in the image thanks to its attention-based design, is constrained by the limited number of samples \cite{vit2}, leading to awful performance. To sum up, conventional vision-only model is ineffective to address ASE dataset. Due to mainstream defective classification methods are CNN-based, it is necessary to develop the multi-modal-based method to deal with the difficulty. The GradCAM visualization is in Figure \ref{gradcam.png}, as the figure shows, CNN focuses on the middle region but is not sensitive to capture the shape or distribution of pink dot. ViT-based model fails to concentrate on the central area,

\begin{table}[]
\scalebox{0.99}{
\begin{tabular}{l|c|cccc}
\hline
Class           & N($\cdot$)&$\mu_x$&$\mu_y$& $\sigma_x^2$& $\sigma_y^2$ \\ \hline
Type-0 (normal) &     225   &    0.04   &   -0.05    &  3.71 &  3.52 \\ \hline
Type-1 (defective) &     92   &    2.73   &  0.59  &  7.38  & 5.52 \\
Type-2 (defective) &    44    &    6.43   &   -3.21    & 8.27 & 8.63 \\
Type-3 (defective) &   50     &   -1.10    &   0.65    & 8.14 &  6.44 \\
Type-4 (defective) &   44     &    -0.21   &   -0.01    & 9.44 &  8.77 \\ \hline
\end{tabular}
}
\caption{The brief descriptive statistic of the proposed ASE dataset, where $N(\cdot)$ denotes as the number of data for a given class. The statistic are extracted by CnOCR \cite{CnOCR}. The rough image-information are shown in Figure \ref{ASEexample.png}, \ref{asecategory.png}.}
\label{tab:asedataset}
\end{table}

\subsection{Prompting to VLM-LLM for ASE dataset}

LLMs possess the capability to retain long-term memory of data and accommodate extensive textual input through their large token capacity. Additionally, they can engage in high-level decision-making through iterative question-answering processes. These attributes form the primary motivation for leveraging LLMs to enhance defect classification. In contrast, VLMs are limited to receiving input in the form of individual image-text pairs. While they can perform rudimentary visual reasoning based on the textual content, their ability is restricted to providing basic descriptions of images.

Industrial defect classification, medical diagnosis, product identification, and similar datasets often contain textual or numerical records in addition to images, especially datasets like ASE dataset, as shown in Figure \ref{ASEexample.png}. Taking advantage of this, we extract numerical information using OCR and combine with our prior knowledge to serve as input for LLM to accomplish subsequent tasks. Recognizing the superior capability of VLM in simple visual reasoning and image description, we utilize straightforward prompting such as \emph{"Please comprehensively describe the distribution and shape of the image"} to obtain basic textual information. Subsequently, combining the output of LLM with the data extracted through OCR and $g(\cdot)$, where $g(\cdot)$ means data preparation. We employ more complex prompting, as depicted in Figure \ref{model.png}, to perform advanced reasoning. The formula can be written as:

\begin{equation}
\begin{aligned}
 \text{VLM's answer} &=  \text{VLM}(\text{prompt}_\text{VLM},\text{Image})\\
 \text{LLM's answer} &=  \text{LLM}(\text{prompt}_\text{LLM},g(\text{OCR}(\text{Image})))\\
\end{aligned}
\label{easy}
\end{equation}

Thanks to the robust zero-shot recognition abilities of VLMs and LLMs, remarkable results are attainable without additional fine-tuning. Additionally, to mitigate the risk of catastrophic forgetting and maintain zero-shot capabilities, we train two adapters tailored to the ASE dataset.

After the VLM-LLM answering, we use a pre-trained image encoder $f_{\text{image}}(\cdot)$ (Resnet-50 \cite{resnet}) and text encoder $f_{\text{text}}(\cdot)$ (BERT-base \cite{bert}) to get a representation for modality fusion. The encoded representation $\mathbf{Z}$ is defined as the Formula \ref{encode}. we fine-tune on the ASE dataset with 30 epochs for warming up those three encoders before alignment and fusion (refer to Section 3-3, 3-4). 


\begin{equation}
\begin{aligned}
\mathbf{Z}_{\text{encoded}}^{\text{Image}} &= f_{\text{image}}(\text{Image})\\
\mathbf{Z}_{\text{encoded}}^{\text{VLM}} &= f_{\text{text}}^{\text{VLM}} (\text{VLM's answer})\\
\mathbf{Z}_{\text{encoded}}^{\text{LLM}} &= f_{\text{text}}^{\text{LLM}}(\text{LLM's answer})\\
\end{aligned}
\label{encode}
\end{equation}

\subsection{Progressive Feature Alignment}

\begin{figure}
    \begin{center}
    \includegraphics[width=0.44\textwidth]{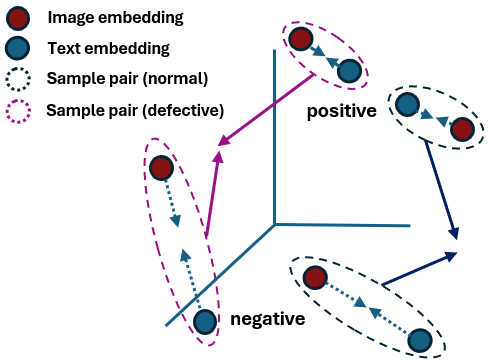}
    \end{center}
    \caption{The encoded feature vectors in low-dimensional space. Our PFA considers the self-similarity within every image-text pairs among all training dataset. By ranking their similarity, low self-similarity data pairs will be regarded as negative samples and added to ${D}_{train}$ with priority \cite{robinson2021contrastive}. It aims to early align negative samples at first to alleviate the difficulty of convergence during alignment.}
     \label{sample.png}
\end{figure}

\begin{figure}
    \begin{center}
    \includegraphics[width=0.48\textwidth]{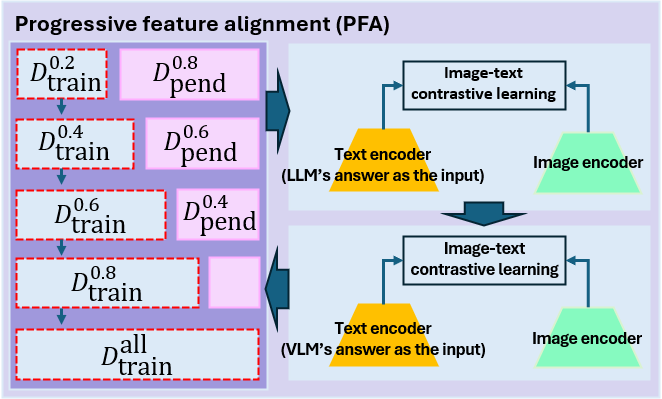}
    \end{center}
    \caption{The pipeline of proposed PFA. In order to handle the problem of insufficient sample-size, it gradually enlarges the training dataset, not trains all data at once.}
     \label{alignment.png}
\end{figure}

At recent years, progressive training strategy (PTS) \cite{progressive} have shown the potential to improve model performance, enhancing the representation understanding without sample-size burden. PTS begins by selecting a subset of positive and negative samples from the training dataset ${D}_{train}$, based on the given initial sampling rate. In our work, we define the positive and negative sample by ranking self-similarity within image-text pairs among training set, negative (low self-similarity) sample will be given priority in the early stages (refer to Figure \ref{sample.png}). PTS divides the entire training set into sub-blocks, trains only a small portion of the whole dataset, and gradually increases the training data with each stage. The remaining samples are used for validation, denoted as ${D}_{pend}$. Benefited by PTS, the network' parameters can be greatly initial and easy to converge. Inspired by \cite{ALBEF,linmultimodality,robinson2021contrastive}, which aims to learn a powerful unimodal representation before fusion, we also use the contrastive learning manner for our model, as shown in Figure \ref{alignment.png}. 

\emph{Directly aligning features is not effective when there are insufficient samples \cite{xu2021dmf}.} The philosophy behind the design of Progressive Feature Alignment (PFA) is that leveraging the PTS's advantage to deal with the inefficacy of multi-modality learning, resulted from the insufficient-large dataset. In proposed PFA block, we use PTS to gradually align different modal representation encoded by VLM-LLM and image encoder, represented as $\mathbf{Z}_{\text{encoded}}^{\text{VLM}}$, $\mathbf{Z}_{\text{encoded}}^{\text{LLM}}$ and $\mathbf{Z}_{\text{encoded}}^{\text{Image}}$ respectively, where all of representation vectors are encoded to lower-dimensional (256-d) representations. For the every single-batch $B$ at the sub-training set at PTS framework, we can formulate the two data-pairs, $(\mathbf{I},T_{\text{LLM}})$ and $(\mathbf{I},T_{\text{VLM}})$ as the input of FPA block. We first align the feature representation ($\mathbf{Z}_{\text{encoded}}^{\text{Image}}$,$\mathbf{Z}_{\text{encoded}}^{\text{LLM}}$) at divided training set, then aligh ($\mathbf{Z}_{\text{encoded}}^{\text{Image}}$,$\mathbf{Z}_{\text{encoded}}^{\text{VLM}}$). Subsequently, we train for 15 epochs and add more training data progressively, until all training data are used. 

For each image and text, we calculate the softmax-normalized image-to-text and text-to-image
similarity as:

\begin{equation}
\begin{aligned}
p^{i2t}_b (I) = \frac{\exp(s(I, T_b)/\tau)}{\sum_{b=1}^{B} \exp(s(I, T_b)/\tau)}\\  p^{t2i}_b (T) = \frac{\exp(s(T, I_b)/\tau)}{\sum_{b=1}^{B} \exp(s(T, I_b)/\tau)}
\end{aligned}
\end{equation}

where $\tau$ is a learnable parameter and $s(\cdot)$ denote as cosine similarity. Finally, the image-text contrastive loss of proposed PFA is defined as the cross-entropy function $H(\cdot)$ between the similarity $p$ and the ground-truth $y$:
{
\small
\begin{equation}
\begin{aligned}
L_{itc}^\text{llm} = \frac{1}{2} \mathbb{E}\left[ H(y^{i2t}(I), p^{i2t}(I)) + H(y^{t2i}(T_\text{llm}), p^{t2i}(T_\text{llm})) \right] \\
L_{itc}^\text{vlm} = \frac{1}{2} \mathbb{E}\left[ H(y^{i2t}(I), p^{i2t}(I)) + H(y^{t2i}(T_\text{vlm}), p^{t2i}(T_\text{vlm})) \right] 
\end{aligned}
\end{equation}
}

\begin{equation}
\begin{aligned}
L_{itc}^\text{total} = L_{itc}^\text{llm} + L_{itc}^\text{vlm}
\end{aligned}
\end{equation}

\subsection{Cross-modality Attention Fusion}

\begin{figure}
    \begin{center}
    \includegraphics[width=0.48\textwidth]{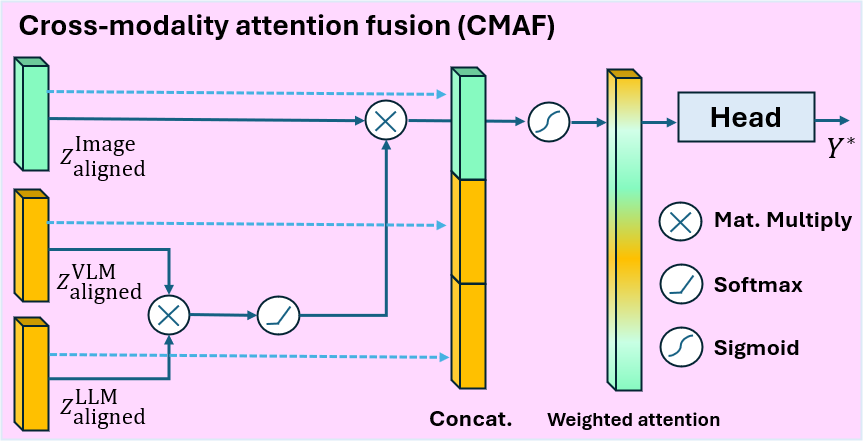}
    \end{center}
    \caption{The proposed CMAF module. It considers the weight of different modalities and fuses these representation adaptively.}
     \label{cmf.png}
\end{figure}

Directly concatenating the representation from different branches may loss some information, leading to reduced performance. Therefore, we proposed cross-modality attention fusion (CMAF) module to incorporate these altogether, avoiding the information missing. Suppose that the aligned features of $f_{\text{text}}(\text{VLM's answer})$, $f_{\text{text}}(\text{LLM's answer})$, and $f_{\text{image}}(\text{Image})$ denoted by $\mathbf{Q}$, $\mathbf{K}$, $\mathbf{V}$, the projection is defined as follows:

\begin{equation}
\begin{aligned}
\mathbf{Q}_i &= \text{Proj.}(\mathbf{Q}) &= \mathbf{Z}_{\text{aligned}}^{\text{VLM}}\\
\mathbf{K}_i &= \text{Proj.}(\mathbf{K}) &= \mathbf{Z}_{\text{aligned}}^{\text{LLM}}\\
\mathbf{V}_i &= \text{Proj.}(\mathbf{V}) &= \mathbf{Z}_{\text{aligned}}^{\text{Image}},
\end{aligned}
\end{equation}
where $\text{Proj.}(\cdot)$ consists of multiple stages to project the input feature into lower dimensional space. First, the $1\times 1$ convolution is used to project the $\mathbf{X} \in \mathbb{R}^{b\times c \times h \times w}$ into $\mathbf{X'} \in \mathbb{R}^{b\times r\times h\times w}$, where $r$ is the reduced number of dimension. We perform the cross-attention \cite{chen2021crossvit} by 

\begin{equation}
    \begin{aligned}
\mathbf{A}_i &= \text{softmax}\left(\frac{\mathbf{Q}_i \cdot \mathbf{K}_i^T}{\sqrt{r}}\right), \\
\mathbf{C} &= \text{Concat.}(\mathbf{A}_0 \cdot \mathbf{V}_0, \mathbf{A}_1 \cdot \mathbf{V}_1, ..., \mathbf{A}_{head} \cdot \mathbf{V}_{head}) , \\
\mathbf{W} &= \text{Sigmoid}(\text{Proj.}(\mathbf{C})) \\
    \end{aligned}
\end{equation}
where $\text{Proj.}_C$ projects the concatenated cross-attentions to get adaptive weights $W$, i.e., $b\times 3\times h\times w$. In this way, we could judiciously fuse the different modalities, the proposed CMAF still remains strong due to its adaptivity, as follows:
\begin{equation}
    \mathbf{Z}_{\text{fused}} = W_1 \cdot \mathbf{Q} + W_2 \cdot \mathbf{K} + W_3 \cdot \mathbf{V}
\end{equation}
where $\mathbf{W}_i$ indicates $i$-th channel of $\mathbf{W}$. 
Eventually, the predicted class $\mathbf{Y}^*$ is obtained via a multi-layer perceptron by 
$\mathbf{Y}^*$ = \text{MLP} ($\mathbf{Z}_{\text{fused}}$).

\subsection{Task-specific Data Augmentation}

Data augmentation is crucial to improve model performance, especially when there is a domain gap or limited sample size. It aims to improve the recognition ability of the model by increasing the diversity of data domain. However, the text and numeric information is paired with image in ASE dataset, common data augmentation strategies, such as geometry and HSV transformation, are incompatible. We design a simple but effective Task-specific data augmentation (TDA), following offline synthesis manner \cite{Shanmugam_2021_ICCV,PreAugNet}, to address the data insufficiency. 

To synthesis the ASE data, we sample data points $(x_i, y_i)$ \emph{N} times from bivariate Gaussian distribution \( \mathcal{N}(\mu_{cls}, \Sigma_{cls}) \) for the different classes.
For each sampled data points, set the pixel at coordinates of sampled point in the image to pink dot. The number of samples \emph{N} obtained depends on the dataset's configuration. For instance, we can utilize OCR-ed data to compute the total points across all radius (refer to Figure \ref{ASEexample.png}). After sampling process, we could obtain the augmented image-text pair data $\mathbf{X}_{aug,cls}^{\text{image}}$ and $\mathbf{X}_{aug,cls}^{\text{text}}$ for ASE dataset. The TDA pipeline can be simply defined as:

\begin{equation}
 \mu_{cls} = [\mu_{x,cls}, \mu_{x,cls}],
 \Sigma_{cls} = \begin{pmatrix} \sigma_{x,cls}^2 & \pm \sigma_{xy,cls} \\ \pm \sigma_{xy,cls} & \sigma_{y,cls}^2 \end{pmatrix}
\end{equation}

\begin{equation}
\mathbf{X}_{aug,cls}^{\text{image}} = \{ (x_i, y_i) \mid (x_{i-1}, y_{i-1}) \sim \mathcal{N}(\mu_{cls}, \Sigma_{cls}), \, i = 1, 2, \ldots, N \}
\end{equation}

\section{Experiment}
\label{sec:experiment}

\begin{table*}[]
\scalebox{0.95}{
\begin{tabular}{l|ccc|cccccc}
 & \multicolumn{3}{c|}{Binary classification} & \multicolumn{6}{c}{Multi-class classification} \\ \hline
Method & \begin{tabular}[c]{@{}c@{}}macro\\ f1-score\end{tabular} & \begin{tabular}[c]{@{}c@{}}f1-score\\ (normal)\end{tabular} & \begin{tabular}[c]{@{}c@{}}f1-score\\ (defective)\end{tabular} & \begin{tabular}[c]{@{}c@{}}macro\\ f1-score\end{tabular} & \begin{tabular}[c]{@{}c@{}}f1-score\\ (normal)\end{tabular} & \multicolumn{1}{c}{\begin{tabular}[c]{@{}c@{}}f1-score\\ (Type-1)\end{tabular}} & \multicolumn{1}{c}{\begin{tabular}[c]{@{}c@{}}f1-score\\ (Type-2)\end{tabular}} & \multicolumn{1}{c}{\begin{tabular}[c]{@{}c@{}}f1-score\\ (Type-3)\end{tabular}} & \multicolumn{1}{c}{\begin{tabular}[c]{@{}c@{}}f1-score\\ (Type-4)\end{tabular}} \\ \hline

DeiT-B \cite{touvron2020training} & 73.59 & 72.72 & 74.46 & 39.96 & 72.88 & 47.82 & 34.09 & 20.00 & 25.00 \\
CrossViT-18 \cite{chen2021crossvit} & 73.16 & 72.39 & 73.91 & 51.33 & 76.44 & 59.78 & 31.81 & 50.00 & 38.63 \\
EfficientNet-b6a \cite{efficientnet} & 77.48 & 74.81 & 80.15 & 64.58 & 83.55 & 71.73 & 56.81 & 54.00 & 56.81 \\
Karmakar et al. \cite{fewshot_defect_1} & 78.02 & 78.07 & 77.97 & 60.69 & 76.44 & 65.21 & 56.81 & 48.54 & 56.54 \\
Xie et al. \cite{ocr3} & 78.46 & 78.41 & 78.50 & 58.78 & 77.77 & 57.60 & 50.00 & 54.00 & 54.54 \\ \hline
\textbf{Proposed} & \textbf{85.65} & \textbf{84.70} & \textbf{86.59} & \textbf{75.91} & \textbf{87.44} & \multicolumn{1}{c}{\textbf{70.65}} & \multicolumn{1}{c}{\textbf{70.45}} & \multicolumn{1}{c}{\textbf{76.00}} & \multicolumn{1}{c}{\textbf{75.00}} \\
\textbf{Proposed with TDA} & \textbf{92.52} & \textbf{92.37} & \textbf{92.67} & \textbf{80.77} & \textbf{91.55} & \multicolumn{1}{c}{\textbf{88.04}} & \multicolumn{1}{c}{\textbf{75.00}} & \multicolumn{1}{c}{\textbf{72.00}} & \multicolumn{1}{c}{\textbf{77.27}}
\end{tabular}
}

\caption{Performance comparison among several defect classification methods for binary/multi-class classification against ASE dataset.}
\label{tab:results}
\end{table*}

\subsection{Implementation Details}
  \textbf{General settings.} 
  The details of ASE dataset have been described in Section 3-1. It contains 455 samples, 325 of which are selected as the training set and 130 as the testing set in our experiments. We extracted external modal information utilizing CnOCR \cite{CnOCR,CTPN}, an efficient OCR package. Instruct-BLIP2 \cite{InstructBLIP2}, a powerful VLM renowned for its adeptness in generating long-context and exhibiting promising performance across several VQA benchmarks, is employed. LLaMa2-7B \cite{llama}, is chosen for generating high-low-level context about decision making.  We use the proposed TDA to synthesis the training sample, from 325 to 650. Both VLM and LLM employed fixed prompting, as detailed in Section 3-2. For the vision encoder, ResNet50 \cite{resnet} is employed. On the other hand, the text encoder utilized BERT-base \cite{bert} following inference from VLM and LLM. Our experiments is on NVIDIA GeForce RTX 3090.

\textbf{Loss function and Hyperparameter-settings.} As for alignment phase, the loss function was described in Section 3-3. After PFA, the loss function is changed to Cross-Entropy for training remains network. The AdamW optimizer \cite{adamw,loshchilov2018decoupled} is employed, where ($\beta_1$, $\beta_2$) are (0.9, 0.999), and the epsilon $\epsilon$ is $1\textrm{e}\!-\!8$. The initial learning rate is $1\textrm{e}\!-\!2$ with the step-wise learning rate decaying every 15 epochs with scaling factor $3/4$, totally trains 60 epochs in fusion phase, and the batch size is $10$.

\begin{figure*}
\includegraphics[width=\textwidth]{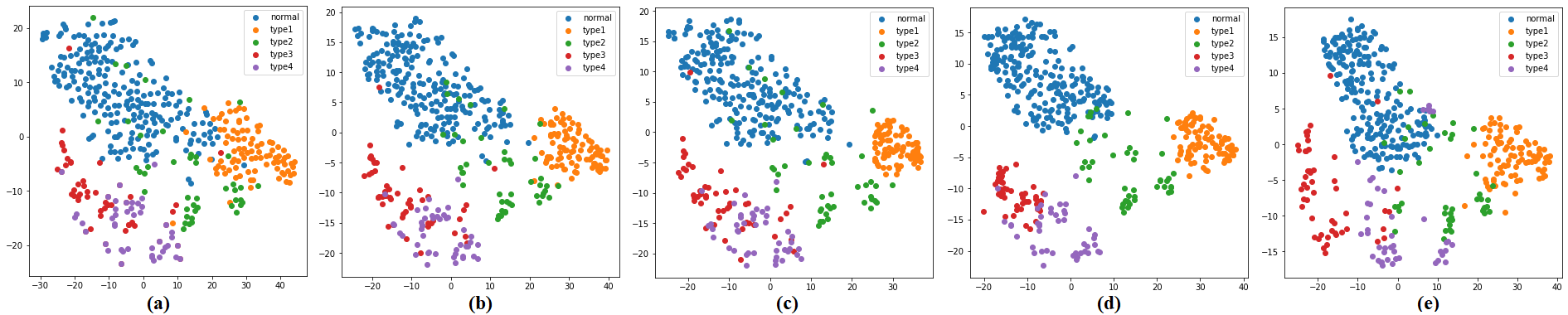}
\caption{The t-SNE visualization of aligned image embedding after proposed Progressive Feature Alignment (PFA) block. (a) un-aligned (b) aligned with 20\% training data (c) aligned with 60\% training data (d) aligned with 100\% training data (e) aligned directly without PTS. The colors blue, orange, green, red, and purple correspond to the classes normal, type-1, type-2, type-3, and type-4 defect, respectively.}
 \label{tsne.png}
\end{figure*}

\textbf{Evaluation metric.} In our experiments, we evaluate model performance using the f1-score for each class and the macro-f1-score. The f1-score is defined as:
\begin{equation}
\text{f1-score} = 2 \times \frac{\text{precision} \times \text{recall}}{\text{precision} + \text{recall}}
\end{equation}
where precision and recall are computed for each individual class. The macro f1-score is the average of the f1-scores for all classes:
\begin{equation}
\text{macro f1-score} = \frac{1}{N} \sum_{i=1}^{N} \text{f1-score}_i
\end{equation}
where $N$ is the number of classes, and $\text{f1-score}_i$ is the f1-score for the $i$-th class. These metrics provide a balanced evaluation of the model's ability to classify each class accurately and its overall performance across all classes.

\subsection{Performance Comparison}

To demonstrate the efficacy of our method in the ASE dataset, we selected three prominent image classification models as baselines. These include CrossViT-18 \cite{chen2021crossvit}, DeiT-B \cite{touvron2020training}, and EfficientNet-b6a \cite{efficientnet}. Additionally, we evaluated our approach against three specialized few-shot-based defect classification frameworks: those proposed by Karmakar et al. \cite{fewshot_defect_1}, which mainly based on transfer learning and ensemble method, Xie et al. \cite{fewshot_defect_2}, which is rooted on exploring extra-feature within given image. 


The experiment results has shown in the Table \ref{tab:results}. Our method demonstrated a outstanding results. Despite being designed to enhance performance through the fusion of multi-scale features and to increase data utilization efficiency, respectively, both CrossViT \cite{chen2021crossvit} and DeiT \cite{touvron2020training} under-perform compared to the simple EfficientNet classifier \cite{efficientnet} on the ASE dataset, whether in binary or multi-class classification tasks. 


\subsection{Ablation Study}

In this section, we ablate important design elements in the proposed method. The results is in the Table \ref{tab:ablation}. In overall, the PFA block is the most beneficial for the model's effectiveness. It aligns the features of AOI images with those of the VLM/LLM image-text pairs, enabling the model to address the minority samples in multi-class issues, especially for the tail classes such as type 2, 3, and 4. Furthermore, our proposed TDA designed for the ASE data also effectively enhances the model's performance.

\begin{table}[]
\scalebox{0.92}{
\begin{tabular}{ccc|c|c}
 PFA & CMAF & TDA & \begin{tabular}[c]{@{}c@{}}macro-f1-score\\ (binary)\end{tabular} & \begin{tabular}[c]{@{}c@{}}macro-f1-score\\ (multi-class)\end{tabular} \\ \hline
 \checkmark &  &  & 83.47 & 69.72 \\
 &\checkmark &  & 78.19 & 64.58 \\
 \checkmark &  \checkmark &  & 85.65 & 75.91 \\
  & \checkmark & \checkmark & 80.60 & 69.19 \\
 \checkmark &  & \checkmark & 86.50 & 76.64 \\\hline
 \checkmark & \checkmark & \checkmark & 92.52 & 80.77
\end{tabular}
}
\caption{Ablation study of proposed method.}
\label{tab:ablation}
\end{table}

\begin{table}[]
\scalebox{0.92}{
\begin{tabular}{c|c|c}
Method & \begin{tabular}[c]{@{}c@{}}macro-f1-score\\ (binary)\end{tabular} & \begin{tabular}[c]{@{}c@{}}macro-f1-score\\ (multi-class)\end{tabular} \\ \hline
Without alignment & 80.60 & 64.19 \\
Direct alignment (whole) & 84.76 & 70.45 \\
PFA (0.2/0.6/1) & 90.46 & 78.13 \\
PFA (0.2/0.4/0.6/0.8/1) & 92.52 & 80.77
\end{tabular}
}
\caption{Ablation study of PFA with different stride.}
\label{tab:pfa}
\end{table}

\begin{table}[]
\scalebox{0.95}{
\begin{tabular}{c|c|c}
Method & \begin{tabular}[c]{@{}c@{}}macro-f1-score\\ (binary)\end{tabular} & \begin{tabular}[c]{@{}c@{}}macro-f1-score\\ (multi-class)\end{tabular} \\ \hline
Direct Concatenation & 86.50 & 76.44 \\
CMAF (w/o. sigmoid) & 87.56 & 77.56 \\
CMAF (w/. sigmoid) & 92.52 & 80.77
\end{tabular}
}
\caption{Ablation study of feature fusion strategy.}
\label{tab:cmaf}
\end{table}

\textbf{Stride settings in PFA block.} The stride is crucial as it pertains to the proportion of data added to the progressive training with each iteration. A smaller stride can more precisely align the embedded vectors between image-text pairs in negative samples, yielding more desirable performance. However, it may lead to additional training duration and computational load. Conversely, a too large stride may result in limited effectiveness, as shown in Figure \ref{tsne.png}. An appropriate stride can guide the model to effectively align with challenging samples without compromising performance. Ablation studies measuring performance across different strides are presented in Table \ref{tab:pfa}.

\textbf{Feature Fusion Strategy.} The method of feature fusion may affect whether certain semantic information can interact well during the training process. Direct concatenation without the use of additional fusion networks, such as self-attention mechanisms, may lead to information loss. In contrast, our proposed CMAF contains the sigmoid function, allowing features from different modalities to adaptively update each other's weights without overly relying on any single modality. The experimental results demonstrating this, as indicated in Table \ref{tab:cmaf}, affirm the efficacy of our approach.

\section{Limitation}
Our proposed method has demonstrated excellent performance on the ASE dataset. Scenarios that may face similar challenges, including data insufficiency and low-quality recorded images, could be (1) defect recognition in industry, (2) product descriptions or classification, and (3) healthcare and medical analysis. While visual modality data is often readily accessible due to the low cost of cameras, paired additional modality data requires databases for recording. This specificity limits the \textbf{generalibility} of the proposed method.

Furthermore, we believe that the proposed PFA is a promising training strategy, especially in scenarios where data scarcity and high inter/intra variability within datasets make model convergence challenging. This method can reduce the difficulty of achieving model convergence or meeting certain criteria during back propagation. However, a drawback is the increased \textbf{computational burden}, resulted from necesarity of holding on the loaded data in memory for every iteration.
\label{sec:discussion}

\section{Conclusion}

\label{sec:conclusion}
In this paper, the special ASE dataset is proposed. The ASE dataset is faced with the issues from data insufficiency and monotonic pattern on image, leading to limitation of conventional deep model like CNN and Transformer. To deal with the challenges mentioned above, we leverages the zero-shot learning capabilities of VLM-LLM to enhance performance on both binary and multi-classification tasks by capturing external-modal features through prompting engineering. Subsequently, the novel progressive feature alignment block, which utilizes progressive training strategy and contrastive learning, effectively aligns image-text representations and progressively incorporates more training data to address the difficulty in alignment presented by a limited sample size. Lastly, the cross-modality attention fusion module adaptively fuses features from different modalities. In summary, our proposed method significantly bolster the model's performance on the ASE dataset, as demonstrated by our experimental results.
{
    \small
    \bibliographystyle{ieeenat_fullname}
    \bibliography{main}

\begin{thebibliography}{74}
\providecommand{\natexlab}[1]{#1}
\providecommand{\url}[1]{\texttt{#1}}
\expandafter\ifx\csname urlstyle\endcsname\relax
  \providecommand{\doi}[1]{doi: #1}\else
  \providecommand{\doi}{doi: \begingroup \urlstyle{rm}\Url}\fi

\bibitem[CnO()]{CnOCR}
{CnOCR}.
\newblock \url{https://github.com/breezedeus/cnocr}.

\bibitem[Ahuja et~al.(2017)Ahuja, Morency, et~al.]{surfey3}
Chaitanya Ahuja, Louis~Philippe Morency, et~al.
\newblock Multimodal machine learning: A survey and taxonomy.
\newblock \emph{IEEE Transactions of Pattern Analysis and Machine Intelligence (TPAMI)}, pages 1--20, 2017.

\bibitem[Akcay et~al.(2022)Akcay, Ameln, Vaidya, Lakshmanan, Ahuja, and Genc]{Anomalib}
Samet Akcay, Dick Ameln, Ashwin Vaidya, Barath Lakshmanan, Nilesh Ahuja, and Utku Genc.
\newblock Anomalib: A deep learning library for anomaly detection.
\newblock \emph{arXiv preprint}, arXiv:2202.08341, 2022.

\bibitem[Bayoudh et~al.(2022)Bayoudh, Knani, Hamdaoui, and Mtibaa]{surfey4}
Khaled Bayoudh, Raja Knani, Fay{\c{c}}al Hamdaoui, and Abdellatif Mtibaa.
\newblock A survey on deep multimodal learning for computer vision: advances, trends, applications, and datasets.
\newblock \emph{The Visual Computer}, 38\penalty0 (8):\penalty0 2939--2970, 2022.

\bibitem[Bhatt et~al.(2021)Bhatt, Malhan, Rajendran, Shah, Thakar, Yoon, and Gupta]{traditional2}
Prahar~M. Bhatt, Rishi~K. Malhan, Pradeep Rajendran, Brual~C. Shah, Shantanu Thakar, Yeo~Jung Yoon, and Satyandra~K. Gupta.
\newblock Image-based surface defect detection using deep learning: A review.
\newblock \emph{Journal of Computing and Information Science in Engineering}, 2021.

\bibitem[Chattopadhay et~al.(2018)Chattopadhay, Sarkar, Howlader, and Balasubramanian]{8354201}
Aditya Chattopadhay, Anirban Sarkar, Prantik Howlader, and Vineeth~N Balasubramanian.
\newblock Grad-cam++: Generalized gradient-based visual explanations for deep convolutional networks.
\newblock In \emph{Proceedings of the IEEE Winter Conference on Applications of Computer Vision (WACV)}, pages 839--847, 2018.

\bibitem[Chenm et~al.(2019)Chenm, Negar, Boris, and Pedro]{Adaptive}
Xing Chenm, Rostamzadeh Negar, Oreshkin Boris, and Pinheiro Pedro.
\newblock Adaptive cross-modal few-shot learning.
\newblock In \emph{Advances in neural information processing systems (NeurlPS)}, 2019.

\bibitem[Chitta et~al.(2023)Chitta, Prakash, Jaeger, Yu, Renz, and Geiger]{Chitta2023PAMI}
Kashyap Chitta, Aditya Prakash, Bernhard Jaeger, Zehao Yu, Katrin Renz, and Andreas Geiger.
\newblock Transfuser: Imitation with transformer-based sensor fusion for autonomous driving.
\newblock \emph{Pattern Analysis and Machine Intelligence (TPAMI)}, 2023.

\bibitem[Chun-Fu et~al.(2021)Chun-Fu, Quanfu, and Rameswar]{chen2021crossvit}
(Richard)~Chen Chun-Fu, Fan Quanfu, and Panda Rameswar.
\newblock Crossvit: Cross-attention multi-scale vision transformer for image classification.
\newblock In \emph{Proceedings of the IEEE/CVF International Conference on Computer Vision (ICCV)}, 2021.

\bibitem[Cootes(2021)]{ocr3}
Xinghui Dong; Christopher J. Taylor; Tim~F. Cootes.
\newblock Defect classification and detection using a multitask deep one-class cnn.
\newblock \emph{IEEE Transactions on Automation Science and Engineering (TASE)}, 2021.

\bibitem[Dai et~al.(2023)Dai, Li, Li, Tiong, Zhao, Wang, Li, Fung, and Hoi]{InstructBLIP2}
Wenliang Dai, Junnan Li, Dongxu Li, Anthony Meng~Huat Tiong, Junqi Zhao, Weisheng Wang, Boyang Li, Pascale Fung, and Steven Hoi.
\newblock Blip-2: Bootstrapping language-image pre-training with frozen image encoders and large language models.
\newblock \emph{arXiv preprint}, arXiv:2305.06500, 2023.

\bibitem[Domen et~al.(2020)Domen, Samo, Jure, and Danijel]{paperwithcodesota2}
Tabernik Domen, Šela Samo, Skvarč Jure, and Skočaj Danijel.
\newblock Segmentation-based deep-learning approach for surface-defect detection.
\newblock 2020.

\bibitem[Dongxu~Guo(2022)]{hf}
Alexandre~Alahi Dongxu~Guo, Taylor~Mordan.
\newblock Pedestrian stop and go forecasting with hybrid feature fusion.
\newblock In \emph{Proceedings of the International Conference on Robotics and Automation (ICRA)}, 2022.

\bibitem[Dylan~Auty(2023)]{adapt2}
Krystian~Mikolajczyk Dylan~Auty.
\newblock Learning to prompt clip for monocular depth estimation: Exploring the limits of human language.
\newblock In \emph{Proceedings of the IEEE/CVF International Conference on Computer Vision (ICCV)}, 2023.

\bibitem[Elhafsi et~al.(2023)Elhafsi, Sinha, Agia, Schmerling, Nesnas, and Pavone]{D-LLM}
Amine Elhafsi, Rohan Sinha, Christopher Agia, Edward Schmerling, Issa A.~D. Nesnas, and Marco Pavone.
\newblock Semantic anomaly detection with large language models.
\newblock \emph{Auton. Robots}, page 1035–1055, 2023.

\bibitem[et~al.(2020)]{gpt3}
Brown et al.
\newblock Language models are few-shot learners.
\newblock In \emph{Advances in Neural Information Processing Systems (NeurIPS)}, pages 1877--1901, 2020.

\bibitem[et~al.(2022)]{flamingo}
Jean~Baptiste et al.
\newblock Flamingo: a visual language model for few-shot learning.
\newblock In \emph{Advances in neural information processing systems (NeurlPS)}, 2022.

\bibitem[et~al.(2023{\natexlab{a}})]{progressive}
Keyao~Wang et al.
\newblock Dynamic feature queue for surveillance face anti-spoofing via progressive training.
\newblock In \emph{Proceedings of the IEEE/CVF Conference on Computer Vision and Pattern Recognition (CVPR)}, 2023{\natexlab{a}}.

\bibitem[et~al.(2019)]{detect_survey_3}
Martin~Mundt et al.
\newblock Meta-learning convolutional neural architectures for multi-target concrete defect classification with the concrete defect bridge image dataset.
\newblock In \emph{Proceedings of the IEEE/CVF Conference on Computer Vision and Pattern Recognition (CVPR)}, 2019.

\bibitem[et~al.(2023{\natexlab{b}})]{llama2}
Touvron et al.
\newblock Llama 2: Open foundation and fine-tuned chat models.
\newblock \emph{arXiv preprint}, arXiv:2307.09288, 2023{\natexlab{b}}.

\bibitem[et~al.(2023{\natexlab{c}})]{fewshot_defect_3}
Xian-Yeow~Lee et al.
\newblock Xdnet: A few-shot meta-learning approach for cross-domain visual inspection.
\newblock In \emph{Proceedings of the IEEE/CVF Conference on Computer Vision and Pattern Recognition Workshop (CVPRW)}, 2023{\natexlab{c}}.

\bibitem[Farady et~al.(2023)Farady, Lin, and Chang]{PreAugNet}
Isack Farady, {Chih Yang} Lin, and {Ming Ching} Chang.
\newblock Preaugnet: improve data augmentation for industrial defect classification with small-scale training data.
\newblock \emph{Journal of Intelligent Manufacturing}, 2023.

\bibitem[Gao et~al.(2020)Gao, Li, Chen, and Zhang]{surfey2}
Jingm Gao, Pengm Li, Zhikuim Chen, and Jianingm Zhang.
\newblock A survey on deep learning for multimodal data fusion.
\newblock \emph{Neural Computation}, 2020.

\bibitem[Haurum and Moeslund(2021)]{Haurum_2021_CVPR}
Joakim~Bruslund Haurum and Thomas~B. Moeslund.
\newblock Sewer-ml: A multi-label sewer defect classification dataset and benchmark.
\newblock In \emph{Proceedings of the IEEE/CVF Conference on Computer Vision and Pattern Recognition (CVPR)}, pages 13456--13467, 2021.

\bibitem[He et~al.(2016)He, Zhang, Ren, and Sun]{resnet}
Kaiming He, Xiangyu Zhang, Shaoqing Ren, and Jian Sun.
\newblock Deep residual learning for image recognition.
\newblock In \emph{Proceedings of the IEEE/CVF Conference on Computer Vision and Pattern Recognition (CVPR)}, 2016.

\bibitem[Hsu et~al.(2023)Hsu, Lee, Hou, and Tsai]{gbacm}
Chih-Chung Hsu, Chia-Ming Lee, Xiu-Yu Hou, and Chi-Han Tsai.
\newblock Gradient boost tree network based on extensive feature analysis for popularity prediction of social posts.
\newblock In \emph{Proceedings of the 31st ACM International Conference on Multimedia (ACMMM)}, page 9451–9455, 2023.

\bibitem[Hugo et~al.(2023)Hugo, Thibaut, Gautier, Xavier, Marie-Anne, Timothée, Baptiste, Naman, Eric, Faisal, Aurelien, Armand, Edouard, and Guillaume]{llama}
Touvron Hugo, Lavril Thibaut, Izacard Gautier, Martinet Xavier, Lachaux Marie-Anne, Lacroix Timothée, Rozière Baptiste, Goyal Naman, Hambro Eric, Azhar Faisal, Rodriguez Aurelien, Joulin Armand, Grave Edouard, and Lample Guillaume.
\newblock Llama: Open and efficient foundation language models.
\newblock \emph{arXiv preprint}, arXiv:2302.13971, 2023.

\bibitem[Jacob et~al.(2018)Jacob, Ming-Wei, Kenton, and Kristina]{bert}
Devlin Jacob, Chang Ming-Wei, Lee Kenton, and Toutanova Kristina.
\newblock Bert: Pre-training of deep bidirectional transformers for language understanding.
\newblock \emph{arXiv preprint}, arXiv:1810.04805, 2018.

\bibitem[Jakob et~al.(2021)Jakob, Domen, and Danijel]{paperwithcodesota1}
Božič Jakob, Tabernik Domen, and Skočaj Danijel.
\newblock Mixed supervision for surface-defect detection: from weakly to fully supervised learning.
\newblock 2021.

\bibitem[Jia et~al.(2022)Jia, Tang, Chen, Cardie, Belongie, Hariharan, and Lim]{jia2022vpt}
Menglin Jia, Luming Tang, Bor-Chun Chen, Claire Cardie, Serge Belongie, Bharath Hariharan, and Ser-Nam Lim.
\newblock Visual prompt tuning.
\newblock In \emph{Proceedings of the European Conference on Computer Vision (ECCV)}, 2022.

\bibitem[Karmakar et~al.(2023)Karmakar, Banerjee, Gidde, Saurav, and Singh]{fewshot_defect_1}
Soumyajit Karmakar, Abeer Banerjee, Prashant Gidde, Sumeet Saurav, and Sanjay Singh.
\newblock Convolutional ensembling based few-shot defect detection technique.
\newblock In \emph{Proceedings of the Thirteenth Indian Conference on Computer Vision, Graphics and Image Processing}, New York, NY, USA, 2023.

\bibitem[khattak et~al.(2023)khattak, Rasheed, Maaz, Khan, and Khan]{MaPLe}
Muhammad~Uzair khattak, Hanoona Rasheed, Muhammad Maaz, Salman Khan, and Fahad~Shahbaz Khan.
\newblock Maple: Multi-modal prompt learning.
\newblock \emph{Proceedings of the IEEE/CVF Conference on Computer Vision and Pattern Recognition (CVPR)}, 2023.

\bibitem[Li et~al.(2021)Li, Selvaraju, Gotmare, Joty, Xiong, and Hoi]{ALBEF}
Junnan Li, Ramprasaath Selvaraju, Akhilesh Gotmare, Shafiq Joty, Caiming Xiong, and Steven Chu~Hong Hoi.
\newblock Align before fuse: Vision and language representation learning with momentum distillation.
\newblock In \emph{Advances in neural information processing systems (NeurIPS)}, pages 9694--9705, 2021.

\bibitem[Li et~al.(2022)Li, Li, Xiong, and Hoi]{BLIP}
Junnan Li, Dongxu Li, Caiming Xiong, and Steven Hoi.
\newblock Blip: Bootstrapping language-image pre-training for unified vision-language understanding and generation.
\newblock In \emph{Proceedings of the International conference on machine learning (ICML)}, pages 12888--12900, 2022.

\bibitem[Li et~al.(2023)Li, Li, Savarese, and Hoi]{BLIP-2}
Junnan Li, Dongxu Li, Silvio Savarese, and Steven Hoi.
\newblock Blip-2: Bootstrapping language-image pre-training with frozen image encoders and large language models.
\newblock \emph{arXiv preprint}, arXiv:2301.12597, 2023.

\bibitem[Liang et~al.(2023)Liang, Li, Zhou, Feng, and Loy]{liang2023iterative}
Zhexin Liang, Chongyi Li, Shangchen Zhou, Ruicheng Feng, and Chen~Change Loy.
\newblock Iterative prompt learning for unsupervised backlit image enhancement.
\newblock In \emph{Proceedings of the IEEE/CVF International Conference on Computer Vision (ICCV)}, pages 8094--8103, 2023.

\bibitem[Lin et~al.(2023)Lin, Yu, Kuang, Pathak, and Ramanan]{linmultimodality}
Zhiqiu Lin, Samuel Yu, Zhiyi Kuang, Deepak Pathak, and Deva Ramanan.
\newblock Multimodality helps unimodality: Cross-modal few-shot learning with multimodal models.
\newblock In \emph{Proceedings of the IEEE/CVF Conference on Computer Vision and Pattern Recognition (CVPR)}, 2023.

\bibitem[Liu et~al.(2023)Liu, Ding, Zhang, and Jiang]{10132374}
Chang Liu, Henghui Ding, Yulun Zhang, and Xudong Jiang.
\newblock Multi-modal mutual attention and iterative interaction for referring image segmentation.
\newblock \emph{IEEE Transactions on Image Processing (TIP)}, 32:\penalty0 3054--3065, 2023.

\bibitem[Loshchilov and H.(2019)]{loshchilov2018decoupled}
Ilya Loshchilov and Frank H.
\newblock Decoupled weight decay regularization.
\newblock In \emph{Proceedings of the International Conference on Learning Representations (ICLR)}, 2019.

\bibitem[Loshchilov and Hutter(2018)]{adamw}
Ilya Loshchilov and Frank Hutter.
\newblock Fixing weight decay regularization in adam.
\newblock In \emph{Proceedings of the International Conference on Learning Representations (ICLR)}, 2018.

\bibitem[Lowe()]{sift}
David~G. Lowe.
\newblock Object recognition from local scale-invariant features.
\newblock In \emph{Proceedings of the IEEE/CVF Conference on Computer Vision and Pattern Recognition Workshop (CVPR)}.

\bibitem[Marco et~al.(2022)Marco, Tom, Bodo, and Bastian]{RudWeh2022}
Rudolph Marco, Wehrbein Tom, Rosenhahn Bodo, and Wandt Bastian.
\newblock Fully convolutional cross-scale-flows for image-based defect detection.
\newblock In \emph{Proceedings of the Winter Conference on Applications of Computer Vision (WACV)}, 2022.

\bibitem[Masoudnia and Ebrahimpour(2014)]{Moe}
S. Masoudnia and R. Ebrahimpour.
\newblock Mixture of experts: a literature survey.
\newblock \emph{Artificial Intelligence Review}, 2014.

\bibitem[Pawłowski~M(2023)]{surfey}
Sysko-Romańczuk~S. Pawłowski~M, Wróblewska~A.
\newblock Effective techniques for multimodal data fusion: A comparative analysis.
\newblock \emph{Sensors (Basel)}, 2023.

\bibitem[Prakash et~al.(2021)Prakash, Chitta, and Geiger]{Prakash2021CVPR}
Aditya Prakash, Kashyap Chitta, and Andreas Geiger.
\newblock Multi-modal fusion transformer for end-to-end autonomous driving.
\newblock In \emph{Proceedings of the IEEE/CVF Conference on Computer Vision and Pattern Recognition (CVPR)}, 2021.

\bibitem[Qian et~al.(2021)Qian, Dawei, Jinxuan, Zhenghao, and Jun]{ocr1}
Xie Qian, Li Dawei, Xu Jinxuan, Yu Zhenghao, and Wang Jun.
\newblock Automatic detection and classification of sewer defects via hierarchical deep learning.
\newblock \emph{IEEE Transactions on Automation Science and Engineering (TASE)}, 2021.

\bibitem[Qu et~al.(2022)Qu, Liu, Wang, and Song]{qu2022transmef}
Linhao Qu, Shaolei Liu, Manning Wang, and Zhijian Song.
\newblock Transmef: A transformer-based multi-exposure image fusion framework using self-supervised multi-task learning.
\newblock In \emph{Proceedings of the AAAI Conference on Artificial Intelligence}, pages 2126--2134, 2022.

\bibitem[Rabiul~Awal(2024)]{vqa}
Aishwarya~Agrawal Rabiul~Awal, Le~Zhang.
\newblock Investigating prompting techniques for zero- and few-shot visual question answering.
\newblock \emph{arXiv preprint}, arXiv:2307.09288, 2024.

\bibitem[Radford et~al.(2021)Radford, Kim, Hallacy, Ramesh, Goh, Agarwal, Sastry, Askell, Mishkin, Clark, Krueger, and Sutskever]{CLIP}
Alec Radford, Jong~Wook Kim, Chris Hallacy, Aditya Ramesh, Gabriel Goh, Sandhini Agarwal, Girish Sastry, Amanda Askell, Pamela Mishkin, Jack Clark, Gretchen Krueger, and Ilya Sutskever.
\newblock Learning transferable visual models from natural language supervision.
\newblock In \emph{Proceedings of the International conference on machine learning (ICML)}, pages 8748--8763, 2021.

\bibitem[Rakshith et~al.(2023)Rakshith, Jayram, Rushil, and Thiagarajan]{CREPE}
Subramanyam Rakshith, T.~S. Jayram, Anirudh Rushil, and Jayaraman~J. Thiagarajan.
\newblock Crepe: Learnable prompting with clip improves visual relationship prediction.
\newblock \emph{arXiv preprint}, arXiv:2307.04838, 2023.

\bibitem[Robinson et~al.(2021)Robinson, Chuang, Sra, and Jegelka]{robinson2021contrastive}
Joshua~David Robinson, Ching-Yao Chuang, Suvrit Sra, and Stefanie Jegelka.
\newblock Contrastive learning with hard negative samples.
\newblock In \emph{International Conference on Learning Representations}, 2021.

\bibitem[Shanmugam et~al.(2021)Shanmugam, Blalock, Balakrishnan, and Guttag]{Shanmugam_2021_ICCV}
Divya Shanmugam, Davis Blalock, Guha Balakrishnan, and John Guttag.
\newblock Better aggregation in test-time augmentation.
\newblock In \emph{Proceedings of the IEEE/CVF International Conference on Computer Vision (ICCV)}, pages 1214--1223, 2021.

\bibitem[Simonyan and Zisserman(2014)]{vgg}
Karen Simonyan and Andrew Zisserman.
\newblock Very deep convolutional networks for large-scale image recognition.
\newblock \emph{arXiv preprint}, arXiv:1409.1556, 2014.

\bibitem[Singh and Desai(2023)]{detect_survey_2}
Swarit~Anand Singh and K.~A. Desai.
\newblock Automated surface defect detection framework using machine vision and convolutional neural networks.
\newblock In \emph{Journal of Intelligent Manufacturing}, 2023.

\bibitem[Syed et~al.(2022)Syed, Hassan, Dang, Suhyeon, Changho, Jaemo, Young-Soo, and Hyeonjoon]{ocr2}
Ibrahim Syed, Minh Hassan, Mehmood Dang, Irfan, Im Suhyeon, Choi Changho, Kang Jaemo, Park Young-Soo, and Moon Hyeonjoon.
\newblock Underground sewer pipe condition assessment based on convolutional neural networks.
\newblock \emph{Automation in Construction}, 2022.

\bibitem[Tabernik et~al.(2020)Tabernik, Šela, Skvarč, and Skočaj]{Surface-Defect-Detection}
Domen Tabernik, Samo Šela, Jure Skvarč, and Danijel Skočaj.
\newblock Segmentation-based deep-learning approach for surface-defect detection.
\newblock pages 759--776, 2020.

\bibitem[Tan and Le(2019)]{efficientnet}
Mingxing Tan and Quoc Le.
\newblock Efficientnet: Rethinking model scaling for convolutional neural networks.
\newblock In \emph{Proceedings of the International conference on machine learning (ICML)}, pages 6105--6114, 2019.

\bibitem[Touvron et~al.(2021)Touvron, Cord, Douze, Massa, Sablayrolles, and Jégou]{touvron2020training}
Hugo Touvron, Matthieu Cord, Matthijs Douze, Francisco Massa, Alexandre Sablayrolles, and Hervé Jégou.
\newblock Training data-efficient image transformers and distillation through attention.
\newblock In \emph{Proceedings of the 38th International Conference on Machine Learning (ICML)}, 2021.

\bibitem[Wang et~al.(2020)Wang, Huang, Sun, Xu, Rong, and Huang]{wang2020cen}
Yikai Wang, Wenbing Huang, Fuchun Sun, Tingyang Xu, Yu Rong, and Junzhou Huang.
\newblock Deep multimodal fusion by channel exchanging.
\newblock In \emph{Advances in Neural Information Processing Systems (NeurIPS)}, 2020.

\bibitem[Wang et~al.(2022)Wang, Sun, Huang, He, and Tao]{wang2022cenpami}
Yikai Wang, Fuchun Sun, Wenbing Huang, Fengxiang He, and Dacheng Tao.
\newblock Channel exchanging networks for multimodal and multitask dense image prediction.
\newblock \emph{IEEE Transaction on Pattern Analysis and Machine Intelligence (TPAMI)}, 2022.

\bibitem[Wenhai et~al.(2023)Wenhai, Zhe, Xiaokang, Jiannan, Xizhou, Gang, Ping, Tong, Jie, Yu, and Jifeng]{visionllm}
Wang Wenhai, Chen Zhe, Chen Xiaokang, Wu Jiannan, Zhu Xizhou, Zeng Gang, Luo Ping, Lu Tong, Zhou Jie, Qiao Yu, and Dai Jifeng.
\newblock Visionllm: Large language model is also an open-ended decoder for vision-centric tasks.
\newblock \emph{arXiv preprint}, arXiv:2305.11175, 2023.

\bibitem[Xu et~al.(2021)Xu, Liu, Zhang, Wang, Li, Huang, Xue, and Fu]{xu2021dmf}
Chengming Xu, Chen Liu, Li Zhang, Chengjie Wang, Jilin Li, Feiyue Huang, Xiangyang Xue, and Yanwei Fu.
\newblock Learning dynamic alignment via meta-filter for few-shot learning.
\newblock In \emph{Proceedings of the IEEE Conference on Computer Vision and Pattern Recognition (CVPR)}, 2021.

\bibitem[Xue and Marculescu(2023)]{xue2022dynamic}
Zihui Xue and Radu Marculescu.
\newblock Dynamic multimodal fusion.
\newblock In \emph{Proceedings of the IEEE/CVF Conference on Computer Vision and Pattern Recognition Workshop (CVPRW), Multi-Modal Learning and Applications Workshop}, 2023.

\bibitem[Xueting et~al.(2024)Xueting, Ce, Yi, Bowen, Ke, and Zhihai]{adapt1}
Hu Xueting, Zhang Ce, Zhang Yi, Hai Bowen, Yu Ke, and He Zhihai.
\newblock Learning to adapt clip for few-shot monocular depth estimation.
\newblock In \emph{Proceedings of the IEEE Winter Conference on Applications of Computer Vision (WACV)}, 2024.

\bibitem[Yajun et~al.(2021)Yajun, Yuanyuan, Fan, Erhu, Zhangnan, and Linhao]{traditional1}
Chen Yajun, Ding Yuanyuan, Zhao Fan, Zhang Erhu, Wu Zhangnan, and Shao Linhao.
\newblock Surface defect detection methods for industrial products: A review.
\newblock \emph{Applied Sciences}, 2021.

\bibitem[Yu et~al.(2024)Yu, Mingzhou, Xiaoqiao, Conghu, and Jing]{fewshot_defect_2}
Gong Yu, Liu Mingzhou, Wang Xiaoqiao, Liu Conghu, and Hu Jing.
\newblock Few-shot defect detection using feature enhancement and image generation for manufacturing quality inspection.
\newblock In \emph{Applied Intelligence}, 2024.

\bibitem[Yuxin et~al.(2023)Yuxin, Wen, Binhui, Quan, Ledell, Xinggang, Tiejun, Xinlong, and Yue]{eva}
Fang Yuxin, Wang Wen, Xie Binhui, Sun Quan, Wu Ledell, Wang Xinggang, Huang Tiejun, Wang Xinlong, and Cao Yue.
\newblock Eva: Exploring the limits of masked visual representation learning at scale.
\newblock In \emph{Proceedings of the IEEE/CVF Conference on Computer Vision and Pattern Recognition (CVPR)}, 2023.

\bibitem[Zhenwei et~al.(2023)Zhenwei, Zhou, Meng, and Jun]{prophet2}
Shao Zhenwei, Yu Zhou, Wang Meng, and Yu Jun.
\newblock Prompting large language models with answer heuristics for knowledge-based visual question answering.
\newblock In \emph{Proceedings of the IEEE/CVF Conference on Computer Vision and Pattern Recognition Workshop (CVPR)}, 2023.

\bibitem[Zhi et~al.(2016)Zhi, Weilin, Tong, Pan, and Yu]{CTPN}
Tian Zhi, Huang Weilin, He Tong, He Pan, and Qiao Yu.
\newblock Detecting text in natural image with connectionist text proposal network.
\newblock 2016.

\bibitem[Zhonghe~Ren and Wu(2022)]{detect_survey}
Ning~Yan Zhonghe~Ren, Fengzhou~Fang and You Wu.
\newblock State of the art in defect detection based on machine vision.
\newblock 2022.

\bibitem[Zhou et~al.(2022{\natexlab{a}})Zhou, Yang, Loy, and Liu]{zhou2022cocoop}
Kaiyang Zhou, Jingkang Yang, Chen~Change Loy, and Ziwei Liu.
\newblock Conditional prompt learning for vision-language models.
\newblock In \emph{Proceedings of the IEEE/CVF Conference on Computer Vision and Pattern Recognition (CVPR)}, 2022{\natexlab{a}}.

\bibitem[Zhou et~al.(2022{\natexlab{b}})Zhou, Yang, Loy, and Liu]{zhou2022coop}
Kaiyang Zhou, Jingkang Yang, Chen~Change Loy, and Ziwei Liu.
\newblock Learning to prompt for vision-language models.
\newblock \emph{International Journal of Computer Vision (IJCV)}, 2022{\natexlab{b}}.

\bibitem[Zhou et~al.(2023)Zhou, Xuecheng, Zhenwei, Meng, and Jun]{prophet1}
Yu Zhou, Ouyang Xuecheng, Shao Zhenwei, Wang Meng, and Yu Jun.
\newblock Prophet: Prompting large language models with complementary answer heuristics for knowledge-based visual question answering.
\newblock In \emph{Proceedings of the IEEE/CVF Conference on Computer Vision and Pattern Recognition Workshop (CVPR)}, 2023.

\bibitem[Zhu et~al.(2023)Zhu, Chen, and Yang]{vit2}
Haoran Zhu, Boyuan Chen, and Carter Yang.
\newblock Understanding why vit trains badly on small datasets: An intuitive perspective.
\newblock \emph{arXiv preprint arXiv:2302.03751}, 2023.

\end{thebibliography}
}


\end{document}